\documentclass[conference]{IEEEtran}

\IEEEoverridecommandlockouts 

\usepackage{cite}
\usepackage{amsmath,amssymb,amsfonts}
\usepackage{array}
\usepackage{graphicx}
\usepackage{textcomp}
\usepackage{hyperref}
\usepackage{xcolor}
\usepackage{algorithmic}
\usepackage{algorithm}
\begin{document}

\title{\LARGE \bf
Vision and Control for Grasping Clear Plastic Bags
}

\author{Joohwan Seo, Jackson Wagner, Anuj Raicura, Jake Kim
\thanks{joohwan\_seo@berkeley.edu, jackson.wagner@berkeley.edu, anj27@berkeley.edu, jake.kim114@berkeley.edu}%
}

\maketitle
\thispagestyle{empty}
\pagestyle{empty}

\begin{abstract}

We develop two novel vision methods for planning effective grasps for clear plastic bags, as well as a control method to enable a Sawyer arm with a parallel gripper to execute the grasps. The first vision method is based on classical image processing and heuristics (e.g., Canny edge detection) to select a grasp target and angle. The second uses a deep-learning model trained on a human-labeled data set to mimic human grasp decisions. A clustering algorithm is used to de-noise the outputs of each vision method. Subsequently, a workspace PD control method is used to execute each grasp. Of the two vision methods, we find the deep-learning based method to be more effective.

\end{abstract}

\section{INTRODUCTION \& RELATED WORK}

State-of-the-art grasping methods in robotics are very sophisticated and robust to many tasks. For example, the paper ``Dex-Net 2.0: Deep Learning to Plan Robust Grasps with Synthetic Point Clouds and Analytic Grasp Metrics" by Mahler et al introduces a grasping method that uses a CNN to evaluate grasp quality and plan an effective grasp given a point cloud and depth image of an object. The method achieves a success rate of 93\% on eight known objects, and achieves 99\% precision on a dataset of 40 novel household objects \cite{c1}. However, even advanced grasping methods like this one struggle with clear plastic bags. 

Clear plastic bags present a number of challenges to robotic systems. First, vision tasks for these objects are exceedingly difficult. The transparent material introduces specularities that depend on lighting, as well as regions that lack reflective signal, both of which confuse sensors and processing systems. Further, clear plastic bags are deformable in unpredictable ways, making it difficult to estimate the quality of a planned grasp before execution. To make matters more challenging, the 2021 paper, ``Modeling, learning, perception, and control methods for deformable object manipulation", by Yin et al., suggests that perception techniques used to determine how an object will deform still lack performance \cite{c5}. 

Nonetheless, picking up clear plastic bags remains a pressing problem in the industry. For example, today in Amazon Warehouses, a significant proportion of the goods purchased by consumers are packaged in clear plastic. This makes the picking and sorting task difficult for robots, so humans do this task in many cases. 

There are several related works that help develop industrial solutions for manipulating plastic bags, but they work with opaque bags and do not address the vision problem associated with clear plastic bags. For example, the 2021 paper ``Initial Results on Grasping and Lifting Deformable Bags with a Bimanual Robot" by Seita et al. found grasping plastic bags at the leftmost and rightmost points yielded a high success rate in the task of grasping and lifting a bag to contain items, suggesting that basic heuristics can be effective in addressing the challenging dynamics of this task \cite{c2}. Further, in the 2023 paper, ``AutoBag: Learning to Open Plastic Bags and Insert Objects" by Chen et al., the researchers propose a self-supervised learning framework where a dual-arm robot learns to recognize the handles of plastic bags using UV-fluorescent markings that are not used at execution time \cite{c3}. Here, this team shows that a deep-learning approach can work for learning to pick effective grasp locations on a plastic bag, but they avoid the additional sensory challenges of clear plastic bags by using opaque plastic bags. 

Other works have indirectly shown that visual recognition of clear plastic objects is achievable with deep learning. For example, the 2019 paper ``Deep Learning Based Robot for Automatically Picking up Garbage on the Grass" by Bai et al. proposes a method capable of recognizing garbage with 95\% accuracy and includes examples of clear bottles properly detected \cite{c4}. 

In our work, we leverage three key insights from these papers: a) simple heuristics can work in the context of picking up bags, b) deep learning can be used to identify good grasp locations on plastic bags and c) deep learning might be able to learn how to decipher images of clear plastic objects. As follows, we develop a classical grasp planning algorithm that uses edge detection and a ``distance from object in bag" heuristic to estimate where contours in clear plastic bags might exist that could be used as grasp targets. We also develop a deep-learning model trained using human-labeled data to process RGB and depth images and return grasp targets and angles. Finally, we develop a de-noising method and a workspace PD controller to execute grasps returned by these vision modules.


\section{METHODS}


\subsection{Problem setup \& criteria for success}
To develop grasping solutions for clear plastic bags, we simplified the problem to that of picking up a ping-pong ball in a large ziplock bag with low-quality parallel grippers. The task is such that the system will need to determine and execute a grasp with the grippers pointing downwards and some rotation around the $z$-axis. Given this setup, we conjecture that approaches developed here can be extended to a variety of objects (heavier, different shapes) in similar packaging with more effective grippers and advanced object detection (in the case of heuristics-based methods). In this test setting, we evaluate grasp procedures by their success rate in picking up the bag as measured empirically through test trials. 

\subsection{System overview}
\begin{figure}[thbp]
\centerline{\includegraphics[width=\linewidth]{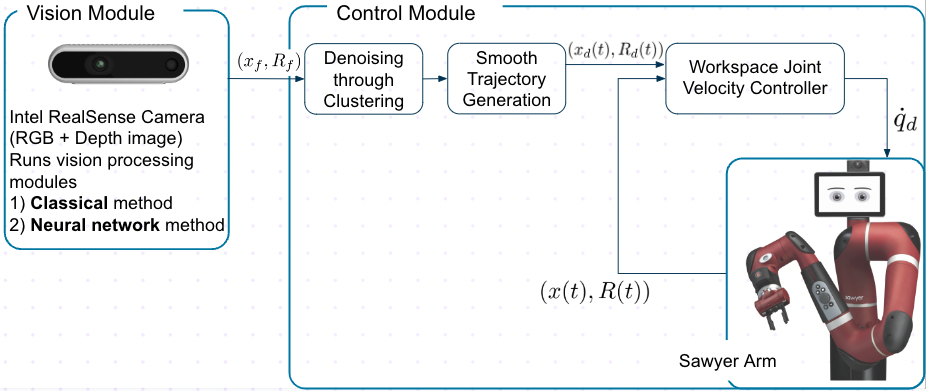}}
\caption{The system consists of a vision module and a control module.}
\label{fig:system}
\end{figure}

Our system consists of both a vision module, for which we implement classical and deep-learning methods and a control module. The vision module takes RGB and depth images from an Intel RealSense camera, and outputs target grasp locations and angles. The control module de-noises a stream of outputs from the vision module via clustering, generates a smooth trajectory, and executes the grasp on the Sawyer arm using a workspace joint velocity controller. See Figure \ref{fig:system} for the system block diagram.

\subsection{Vision module}
We implemented classical and deep-learning-based computer vision modules discussed in the following sections. 

\subsubsection{Classical computer vision module}
The goal of the classical vision module is to find an elevated crease or edge on the plastic bag. To do so, it looks for edges in the RGB image and picks a ``big" one that is somewhat close to the ball (a heuristic for elevation). The precise procedure is as follows:
\begin{itemize}
    \item Take RGB image as input and apply a Gaussian Blur followed by Canny edge detection. Use color thresholding to find the center and radius of the ball.
    \item Find the contours that outline the edges and return them as a list of polygons. 
    \item Find the \textbf{mean} of each polygon and the \textbf{slope} using linear regression. Convert the slope to a $\theta_d$ value.
    \item For each polygon with a perimeter greater than a threshold, find the one with the mean closest to 1.1*radius of the ball away from the center of the ball. 
    \item Scale and shift the associated mean to convert to workspace coordinates and return along with $\theta_d$ as the grasp target and angle. 
\end{itemize}

\begin{figure}[thbp]
\centerline{\includegraphics[scale=0.5]{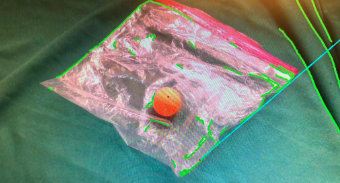}}
\caption{Classical vision module output. The red dot and cyan line at the bottom-middle of the image are the target grasp location and theta respectively. Green polygons are contours around prospective edges.}
\label{fig:classical}
\end{figure}

Find an example output from the classical vision module in Figure \ref{fig:classical}. 

\subsubsection{Deep-learning vision module}
The goal of the deep-learning vision module is to replicate human grasp decisions as represented in a dataset our team constructed. To build the dataset used for training, our team took $413$ RGB + depth images of the bag-ball setup in a variety of configurations, cropped each to leave the center $1/4$th, resized to $36\times 64$, and manually specified target grasp locations and $\theta_d$ based on observation of the real bag and the downsized images. 

\begin{figure}[thbp]
\centerline{\includegraphics[width=\linewidth]{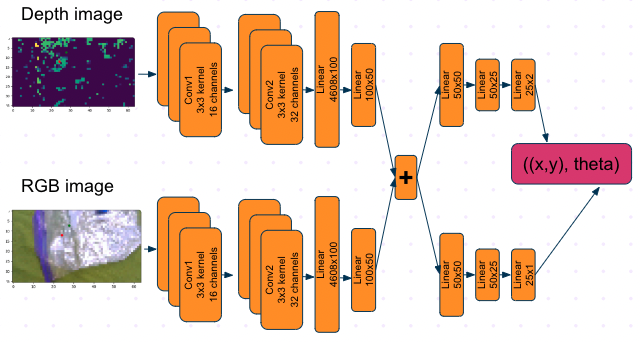}}
\caption{The neural network processes RGB and depth images in parallel before combining embeddings and then splitting again to predict target (x,y) and $\theta$ respectively. }
\label{fig:nn_architecture}
\end{figure}

\begin{figure}[thbp]
\centerline{\includegraphics[width=\linewidth]{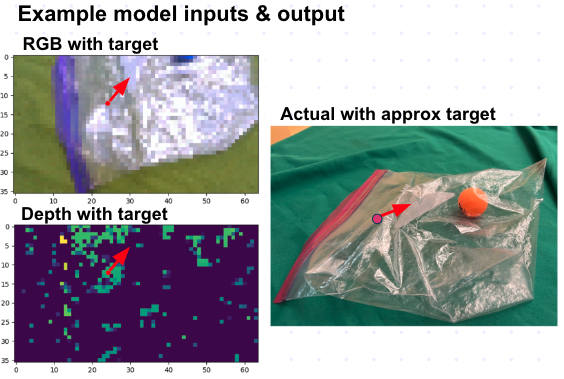}}
\caption{The neural network learned to interpret both the depth and RGB inputs to decide to grasp along a prominently raised crease in the bag in this example.}
\label{fig:nn_in_out}
\end{figure}

The model architecture consists of two parallel convolutional neural networks that process the downsized RGB and depth images for a scene respectively. The resulting embeddings for each of these images are subsequently added. The network then splits into two linear networks, one for predicting the target grasp location, and the other for predicting the grasp $\theta_d$. The model was trained using L1 loss for 50 epochs. See Figure \ref{fig:nn_architecture} for an illustration of the architecture and \ref{fig:nn_in_out} for an illustration of an example model output.

\subsection{Control module}
The control module first de-noises the input via a clustering technique, then generates a smooth trajectory, and finally executes the trajectory and grasp. 
\subsubsection{De-noising via clustering}
Outputs from the vision module can vary significantly due to sensor noise, varying lighting conditions, and the existence of multiple competing viable grasps. To try and pick "the best" grasp returned and make our system more consistent, we apply the following procedure. 
\begin{itemize}
    \item Collect published grasp proposals from the past 10 seconds
    \item Find the distance between all grasp targets and all other grasp targets
    \item Group grasps into clusters based on a distance threshold
    \item Pick the center of the cluster with the most points, and the mode $\theta$ value in that cluster as the desired grasp
\end{itemize}
\subsubsection{Trajectory generation and control}
Given the de-noised desired end-effector pose ($p_f$, $\theta_d$), we first generate the smooth trajectory starting from the current position to the final desired end-effector pose. Note that for the easiness of the problem setup, we always let the end-effector be vertical to the ground, i.e., the end-effector $z$ axis (axis of rotation of the final joint) is always aligned to the spatial $z$ axis. It turns out that when $\theta_d = 0$, the rotation matrix of the end-effector is given by
\begin{align}
    R_c = \begin{bmatrix}
        -1 & 0 & 0 \\ 0 & 1 & 0 \\ 0 & 0 & -1
    \end{bmatrix} \nonumber
\end{align}
The desired rotation matrix $R_f$ is then calculated by rotation matrix around $z$ axis as follows:
\begin{align}
    R_f = R_c R_z(\theta_d) = R_c \begin{bmatrix}
        \cos{\theta_d} & -\sin{\theta_d} & 0 \\
        \sin{\theta_d} & \cos{\theta_d} & 0 \\
        0 & 0 & 1
    \end{bmatrix}
\end{align}

The smooth positional reference trajectory $p_d(t) \in \mathbb{R}^3$ is calculated from the third-order polynomial time-trajectory, using the same method from project 1B \cite{proj1b}, e.g.,
\begin{align}
    p_d(t) = a + bt + ct^2 + dt^3
\end{align}
where $a,b,c,d \in \mathbb{R}^3$ is the parameter needs to be calculated. We denote $a = [a_x, a_y, a_z]^T$ and similarly for $b,c$ and $d$. The constraints for the smooth trajectory is
\begin{align}
    \begin{array}{ll}
    t_i \leq t \leq t_f &\\
    p_d(t_i) = \text{current position}, &  p_d(t_f) = p_f \nonumber \\
    \dot{p}_d(t_i) = 0, & \dot{p}_{d}(t_f) = 0
    \end{array}
\end{align}
The parameters $a,b,c,d$ can be calculated from $Ap_i = q_i$, for $i \in \{x,y,z\}$ where
\begin{align} \label{eq:line_trajectory_matrix}
    A \!=\! \begin{bmatrix}
    1 & t_i & t_i^2 & t_i^3\\
    1 & t_f & t_f^2 & t_f^3\\
    0 & 1  & 2t_i  & 3t_i^2\\
    0 & 1  & 2t_f  & 3t_f^2
    \end{bmatrix}, \;
    p_i \!=\! \begin{bmatrix}
    a_i\\
    b_i\\
    c_i\\
    d_i
    \end{bmatrix}, \;
    q_i \!=\! \begin{bmatrix}
    p_{d,i}(t_i) \\
    p_{d,i}(t_f) \\
    0           \\
    0
    \end{bmatrix} 
\end{align}
The smooth rotational trajectory $R_d(t)$ is parameterized as the polynomial as follows:
\begin{align}
    R_d(t) &= R_e \exp{(\hat{\omega}(t))}, \nonumber \\
    \omega(t) &= a_r + b_r t + c_r t^2 + d_r t^3,
\end{align}
where $R_c$ is current rotational matrix, and $a_r, b_r, c_r, d_r \in \mathbb{R}^3$ are the coefficients to be calculated. 
the following constraints are considered:
\begin{align}
    \begin{array}{ll}
         t_i \leq t \leq t_f &\\ 
         R_d(t_i) = \text{current orientation} (R_e), & R_d(t_f) = R_f\\
         w(t_i)= 0 , & w(t_f) = 0.
    \end{array}
\end{align}
The parameters $a_r = [a_{r,1}, a_{r,2}, a_{r,3}]^T$, $b_r, c_r$ and $d_r$ can be calculated from the following equation:
\begin{align}
    A p_j &= q_j, \text{for} \; j \in \{1,2,3\} \\
    p_j &= [a_{r,j}, b_{r,j}, c_{r,j}, d_{r,j}]^T,  q_j = [w_{init,j}, w_{final,j}, 0, 0]^T  \nonumber
\end{align}
where 
\begin{align}
    w_{init} &= \log{(I_{3,3})}, \quad w_{final} = \log{(R_e^T R_f)} \nonumber
\end{align}
with $w = \log(R)$, is the logarithm mapping given by $\log : SO(3) \to \mathbb{R}^3$, 
\begin{align}
    w = \left(\dfrac{\phi}{2\sin{\phi}} (R - R^T) \right)^\vee,\;\; \phi = \arccos{\dfrac{\text{trace}(R) - 1}{2}}.
\end{align}

We employed Workspace PD velocity control for our control method. In this framework, the desired joint angle velocity $\dot{q}_d$ is considered as the control input. First, the error vector in the Cartesian coordinate $e_p$ is calculated by
\begin{align}
    e_p(t) = p(t) - p_d(t)
\end{align}
where $p$ can be obtained by solving forward-kinematics, and $p_d(t)$ is from smooth trajectory signal. On the rotational counterpart, given the desired rotation matrix $R_d = [r_{d,1}, r_{d,2}, r_{d,3}]$ and current rotation matrix $R_e = [r_{1}, r_2, r_3]$, the rotational error vector $e_o$ is obtained as follows \cite{ochoa2021impedance, shaw2022rmps}:
\begin{align}
    e_o = (r_{d_1}\! \times\! r_1 + r_{d_2}\! \times \! r_2 + r_{d_3}\! \times\! r_3)
\end{align}
Note that this rotational error notation is first utilized in \cite{luh1980resolved}, and is known to be equivalent to the $\sin$ of the angle to rotation $R_d$ to $R_e$ with respect to the axis of rotation. Finally, the total error vector $e$ is obtained as $e = [e_p^T, e_o^T]^T$.

Based on these error signals, the control inputs are calculated as follows:
\begin{align}
    \dot{q}_d = J_s^{\dagger} (-K_p e - K_d \dot{e} + V_d),
\end{align}
where $J_s^\dagger$ is Moore-Penrose pseudo-inverse of the spatial Jacobian matrix, $\dot{e}$ is numerically obtained time-derivative of $e$, and $V_d = [\dot{p}_d^T, \omega(t)^T]^T$ is the feed-forward term. In our experiment, the gain matrices are characterized by the scalar gains, i.e., $K_p = k_p I_{6,6}$, $K_d = k_d I_{6,6}$, with $k_p = 0.8$ and $k_d = 0.4$.
\section{RESULTS}

\subsubsection{Experimental setup}
To test the different versions of our system, we used two different gripper configurations and two different bag configurations. The gripper either had rubber tips, or smooth tape tips. The bag was either placed such that its surface was smooth / flat, or crumpled. For each trial we recorded if the grasp was successful, and if the grasp target was good as evaluated subjectively via observation. For sake of description, a trial would consist of a) novel placement of the bag in either the flat or crumpled configuration, b) execution of the vision and control modules such that the Sawyer arm attempted a grasp, and b) the recording of whether or not the grasp was successful (i.e., the bag was lifted to the Sawyer starting configuration) and if the grasp seemed logical or good from a human perspective. 

\subsection{Classical vision experimental results}
\begin{figure}[thbp]
\centerline{\includegraphics[width=\linewidth]{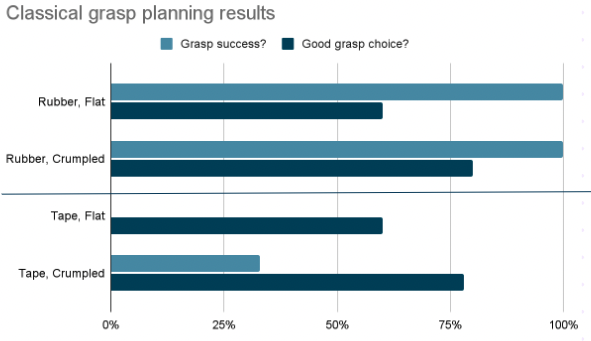}}
\caption{Experimental results with classical vision module. N = 5 for both rubber gripper experiments and Tape, Flat experiment. N = 9 for Tape, Crumpled.}
\label{fig:classical_results}
\end{figure}

When the classical vision module was tested with the rubber gripper, the system successfully grasped the bag 100\% of the time. When the bag was flat, the system made effective grasp choices (e.g., picked the edge of the bag) 3/5 trials, and when the bag was crumpled it picked good grasps 4/5 trials. However, when the same module was tested with the tape gripper, which had significantly less grip than the rubber gripper and therefor required more precise grasp planning, the system succeeded 0\% of the time when the bag was flat and only 3/9 trials when the bag was crumpled, with two trials where the bag slipped out of the grasp on the way up. Good grasp choice rates were very similar with the tap gripper as expected, validating the consistency of our subjective grasp evaluation. The conclusion from these results was that showing proficiency with the tape gripper in the crumpled bag setting would represent a meaningful improvement, so we put our deep-learning vision module to the test in this setting. See Figure \ref{fig:classical_results} for a summary of the results.

\subsection{Deep-learning vision experimental results}
\begin{figure}[thbp]
\centerline{\includegraphics[width= 0.7 \linewidth]{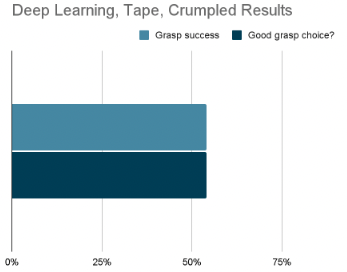}}
\caption{Experimental results with deep-learning vision module. N = 11 for Tape, Crumpled experiment.}
\label{fig:nn_results}
\end{figure}

The deep-learning vision module was evaluated with the tape gripper and crumpled bag configurations. Out of 11 trials, the system executed a successful grasp 6 times, and made a good grasp choice each of those 6 times. However, it is important to note that when the system failed to grasp the bag (and also made a poor grasp choice), it was often very close to a great grasp location, many times even with the correct gripper $\theta$ (e.g., gripper slightly offset from crease in bag, so pushes the crease down). Based on qualitative observation, it seemed clear that the deep-learning vision module was picking grasps that more closely resembled the grasp choices that we would make as humans in both success and failure cases. In success cases, it is also notable that the deep-learning system picked some effective grasps that would not have been chosen by the classical module, as they would not be considered by the rule-based algorithm. See figure \ref{fig:nn_results} for a summary of the results. 

\section{DISCUSSION}
To recap, we evaluated our classical grasp planning method using both rubber and tape grippers, as well as crumpled and flat bags, but found that the rubber gripper would create ridges in the bag as a result of the friction it had on the surface, leading to an inflated success rate. The tape grip showed results more similar to what was expected, and showed limited success in the crumpled case, and no success in the flat case. We saw the deep learning approach generally worked better than the classical method in the tape grip and crumpled bag configuration, picking locations that were at or close to locations where humans would have grasped the bag. It showed a marked improvement in choosing the optimal grasp locations, and this was reflected in the measured success rate.

We found that this problem of grasping transparent deformable objects is indeed not an easy problem to solve. In our experiment, we simplified the problem to a single type of bag with a single object. We then applied deep learning to achieve a marked improvement over the classical technique we developed. However, it is important to note that because of the highly-specific dataset we used and the controlled environment in which we tuned our system, our solution as is cannot be extended to enable grasping of common transparent objects. For the classical model, we were unable to get consistent results even in our simplified ping pong ball and Ziploc system.

The classical method was seen in our experimentation to be vulnerable to changes in the light source, as we could not maintain a true constant and dominating light source in our experimentation setup. The deep learning method showed promise but comes with some issues as well, namely the aforementioned small network size and dataset. Another issue that could arise is operator mismatch in data labeling, because currently, the data had been labeled by multiple different humans with varying ideas of what the optimal grasp position and angle was. Furthermore, even after a grasp location has been decided on by either of the two deduction methods, we are unable to identify and score the proficiency of the grasp aside from whether or not a human would agree with the grasp location. This is once again a qualitative measurement, and we need a more consistent, unbiased metric for quantifying this. In the DexNet paper, a grasp metric was introduced for the scoring, and based on that, the desired grasp points are obtained, which could maximize the grasp metric; however, with deformable objects, the proper grasp technique becomes orders of magnitude greater than rigid \cite{c1}. 

We found that it difficult to find proper heuristics for the classical model to find grasping locations. While we were able to image process the contours and features of the bag to a relatively high degree, we relied on the knowledge that the center of the ping pong ball would have to increase the height of the bag. Using this, we defined a distance metric, forcing the algorithm to choose a location close to the ball. This can definitively be replaced with a more accurate and reliable method that will work for all shapes and sizes of objects in the Ziploc, as our current method will not. We attempted to use the depth readings from our camera to solve this issue, but we found processing and interpreting the data to be quite cumbersome. Given more time, this is something that could be explored further and taken into account in this classical method. Our camera also is able to give point cloud data that we did not look into much over the course of this project, which also could prove to be a valuable parameter when determining grasp locations. 

Another set of difficulties that we encountered was in developing the neural network for our deep learning method. Initially, our NN was much more complex, with three convolutional layers per path with 64, 128, and 256 channels, respectively. Using this in combination with MSE loss outputted grasp locations at the center of the image frame regardless of the state of the bag, which is known as a multi-modality problem. Eventually, we simplified and ended up with the current model as previously described in figure \ref{fig:nn_architecture}, and used an L1 loss function which we found to be much more accurate. Over the course of debugging, we tried implementing batch normalization as well, but this did not seem to positively impact our outcomes. Given more time and neural network expertise, we believe that we could have improved our model at least to a more reliable level. We would also like to experiment with different types of input and output to the NN, as we were only able to go with our initial idea of inputting RGB pixels and depth arrays and outputting $(x, y, \theta)$. We contemplated having the NN output a mask of the grasp location as well, which could be something to explore in the future. Finally, we would have liked to expand on the size of the dataset greatly, something that could be pursued in the future. 

There are several other approaches that could be taken into account as possible future works. 
1) The first approach is that the current deep-learning problem can be formulated as something similar to the {\it behavior cloning} problem. Under that imitation problem setup, we could utilize Dataset Aggregation (DAgger) method which is provided in \cite{ross2011reduction}. By using this method, we could actively aggregate our dataset. However, there might be subtleties in adopting the DAgger method, while the imitation learning framework collects the trajectories, our approach is a one-shot-done task. 
2) The second approach is to make neural network output probabilities (or probability density functions) instead of direct desired points. This approach can be thought about two more branches. 2a) Utilization of Mixture Density Network or Bayesian Neural Network to resolve the multi-modality issue we encountered, where we used clustering to deal with this issue. 2b) Make the neural network return the expected success probabilities for each pixel - something similar to semantic segmentation problems. Some manifest drawbacks of these methods are that a lot larger dataset is required for the probability-based approach to work, to the size where law-of-large numbers or Central Limit Theorem could be applied. 

In terms of hardware, we did not generally explore many of the possibilities. Grasping, however, is both a hardware and software problem. For our test setup, we would have liked to physically fix the camera onto a landmark in the environment rather than just a tripod. We also would have liked to iterate on the design of the gripper more and perhaps pursue having the robot come in from the side of the bag with a horizontal gripper or simply use a vacuum gripper. The ends of the grip also could be coated in silicone, sandpaper, or any other substance with high grip properties, to name a few more ideas in the materials used, rather than just the tape and rubber that we ended up utilizing. Another main hardware issue that could easily be improved is the control of the light source. Our setup did not let us really change the intensity of light nor the direction of light the camera saw, which we hypothesize drastically affected the noise that we saw to hinder our results. Furthermore, another idea is to use a light source with a known pattern to enable more precise mapping of the contours of the bag. 

Overall, we were able to demonstrate some level of feasibility in grasping transparent plastic bags using both classical computer vision techniques as well as deep learning methods. We demonstrate the power of deep-learning methods to address complex problems, as abstracting away sensor data interpretation via learning can improve results as well as decrease implementation time drastically. Our end-to-end learning system provided results that were somewhat similar to our training dataset. Even though our model was overfitted to the dataset, it showed better results than our classical method in the crumpled-bag case.


\section{APPENDIX}
Find the code for our implementation on \href{https://github.com/jaxwagner/eecs206b_final_project}{github}.

Find example demo videos at this \href{https://docs.google.com/presentation/d/1Fo5i_c8PX-CudURtu07XJa0gvtH4dwECKNj1_J_-ux8/edit#slide=id.g23e19717af8_0_381}{link}.

\bibliographystyle{bibliography/IEEEtran}
\bibliography{bibliography/references} 







\end{document}